\documentclass[11pt]{article}
\usepackage{coling2018}
\usepackage{times}
\usepackage{url}
\usepackage{latexsym}
\usepackage{pgfplots}
\usepackage{adjustbox}

\pgfplotsset{compat=1.14}

\usepackage[disable]{todonotes}   

\newcommand{\note}[4][]{\todo[author=#2,color=#3,size=\scriptsize,fancyline,caption={},#1]{#4}} 

\newcommand{\manuel}[2][]{\note[#1]{Manuel}{orange!40}{#2}}

\usepackage{CJKutf8}

\slname{Nahuatl}

\title{Challenges of language technologies for the \\indigenous languages of the Americas}

\sltitle{Masehualtlahtoltecnologias ipan Americatlalli}

\author{Manuel Mager\\
  Instituto de Investigaciones en Matem\'{a}ticas \\
  Aplicadas y en Sistemas \\
  Universidad Nacional Aut\'{o}noma de M\'{e}xico \\
  {\tt mmager@turing.iimas.unam.mx} \\\And
  Ximena Gutierrez-Vasques\\
  GIL IINGEN \\
  Universidad Nacional \\ Aut\'{o}noma de M\'{e}xico \\
  {\tt xim@unam.mx} \\\AND
  Gerardo Sierra\\
  GIL IINGEN \\
  Universidad Nacional \\ Aut\'{o}noma de M\'{e}xico \\
  {\tt gsierram@iingen.unam.mx\ \ } \\\And
  Ivan Meza\\
  Instituto de Investigaciones en Matem\'{a}ticas \\
  Aplicadas y en Sistemas \\
  Universidad Nacional Aut\'{o}noma de M\'{e}xico \\
  {\tt \ \ ivanvladimir@turing.iimas.unam.mx} \\}

\date{}

\begin{document}
\maketitle

\vspace{50px}

\begin{abstract}
Indigenous languages of the American continent are highly diverse. However, they have received little attention from the technological perspective. In this paper, we review the research, the digital resources and the available NLP systems that focus on these languages. We present the main challenges and research questions that arise when distant languages and low-resource scenarios are faced. We would like to encourage NLP research in linguistically rich and diverse areas like the Americas.  
\end{abstract}

\makesltitle
\begin{slabstract}
In nepapan Americatlalli imacehualtlahtol, inin tlahtolli ahmo quinpiah miac tlahtoltecnologías (“tecnologías del lenguaje”). Ipan inin amatl, tictemoah nochin macehualtlahtoltin intequiuh, nochin recursos digitales ihuan nochin tlahtoltecnologías in ye mochiuhqueh. Cequintin problemas monextiah ihcuac tlahtolli quinpiah tepitzin recursos kenin amoxtli, niman, ohuic quinchihuaz tecnología ihuan ohuic quinchihuaz macehualtlahtolmatiliztli. Cenca importante in ocachi ticchihuilizqueh tlahtoltecnologías macehualtlahtolli, niman tipalehuilizqueh ahmo mopolozqueh inin tlahtolli.
\end{slabstract}


\section{Introduction}
\label{introduction:ref}
The American continent is linguistically diverse, it comprises many indigenous languages that are nowadays spoken from North to South America. There is a wide range of linguistic families and they exhibit linguistic phenomena that are different from the most common languages usually studied in Natural Language Processing (NLP).  There are approximately $28$ million\footnote{Each country has its own methodology and criteria to estimate the amount of speakers. This is the sum of all estimations.} people who self identify as members of an indigenous group~\cite{wagner2016lenguas} and they speak around $900$\footnote{This number varies depending on the classification criteria used on different studies.} native or indigenous languages. This represents an important cultural and linguistic richness. This richness was captured by the following quote from \newcite{mcquown1955indigenous} ``in one small portion of the area, in Mexico just north of the Isthmus of Tehuantepec, one finds a diversity of linguistic type hard to match
on an entire continent in the Old World''.
In spite of this, few language technologies have been developed for these languages, moreover, many of the indigenous languages spoken in the Americas face a risk of language extinction.

The aim of this work is to explore the research in the NLP field for the indigenous languages spoken in the American continent and to encourage research for these languages. We stress the need of developing language resources and NLP tools for these languages and we point out some of the challenges that arise when working on this field. Since indigenous languages are digitally scarce, developing technologies can have a positive social impact for the communities which depend on these languages. The great diversity of these languages posses interesting scientific challenges, e.g., adapting well established approaches, creation of new methods, collecting new data. Tackling these challenges contributes to building more general computational models of language, and to get a deeper insight into human language understanding. Moreover, many statistical NLP methods seek to achieve language independence, however they often lack of linguistic knowledge or they do not cover the broad diversity of languages \cite{bender2011achieving}. In this sense, it is important to acknowledge the characteristics of the indigenous languages of the Americas as a way of complementing the current NLP methods.

\paragraph{Contributions.} To sum up, we made the following contributions: (i) we gave a brief introduction into the diverse language space of the indigenous languages of the American continent; (ii) we provide an overview of the existing digital corpora and language sytems that have been developed for some of the languages spoken in the Americas; (iii) and we discusses the advances, used methodologies, challenges and open questions for the most researched NLP tasks in these languages.

In order to maintain the information of this paper updated of the computer readable resources, developed systems and scientific research we made a public available list\footnote{Updated list of resources for Indigenous languages of the Americas: \url{https://github.com/pywirrarika/naki}} with the latest advances for the Indigenous Languages of the Americas.  

\subsection{Languages overview}
Linguistic typology is a field that studies the different languages of the world and tries to establish the relationships among them. This is not an easy task since there is still limited knowledge about many languages, specially in highly diverse regions. 
However, according to several linguistic atlas\footnote{Based on Ethnologue~\cite{simons2017ethnologue}, Glottolog\cite{nordhoff2013glottolog}, WALS \cite{wals}.}, 
there are around 140 linguistic families in the world, from these linguistic families almost $40\%$ are native to the Americas. Nowadays, approximately $900$ different indigenous languages are spoken of this region, making this continent a linguistically diverse territory.

Americas native languages can exhibit very different linguistic phenomena, these typological features are important to be taken into account when developing language technologies \cite{o2016survey}. It would be hard to provide a general description of all the languages spoken in the Americas, however, we would like to highlight some of the linguistic features that usually represent a challenge when doing NLP.

Many languages in North America tend to have a high degree of morphological synthesis, i.e., many morphemes per word~\cite{mithunpolysynthesis}. For instance, languages that belong to the Eskimo-Aleut family (native to Canada, Alaska, Greenland and Siberia) are highly polysynthetic suffixing languages. Languages from  other linguistic families spoken in North America also show specific degrees of agglutination, polysynthesis, and they have morphemes that express a wide range of functions and nuances of space or direction~\cite{mithun2001languages}. Languages with this type of phenomena usually have compact word constructions that are equivalent to whole sentences in other languages like English.

Another linguistic phenomena that is found in many languages spoken in the Americas is the tone, i.e., languages where the pitch is important to distinguish one word from another (the tone can express lexical meaning and grammatical function). Some linguistics families like Oto-manguean (spoken in Mexico) have languages with many types of tones. The orthography of these languages need to mark the wide range of tones, however, many indigenous languages face a lack of orthographic normalization. This can be specially problematic when trying to do NLP and processing text documents.

In general, the native languages of the Americas have a limited digital text production, in some cases they may have a strong oral tradition but not a written one. Due to social and political reasons, alphabetization and education programs are not always available for native speakers. Details about the languages families for which we were able to identify digital and technological resources, are given in Appendix A

\section{Corpus and digital resources}
\label{corpora:ref}

Most of the current state of the art methods in NLP are data-driven approaches that require vast amounts of corpora in order to achieve good performance. Widely popular machine learning methods and vector space representations, e.g., neural networks and word embeddings, often rely in big monolingual corpora.

Annotated and unannotated corpora are required for several NLP tasks. For instance, parallel corpora are essential for building statistical machine translation (MT) systems, while morphological annotated data is essential for POS (Part-of-Speech) taggers and morphological analyzers, just to name a few.

In the case of MT, the most common sources for gathering large amounts of parallel data include specific domain texts such as parliamentary proceedings, religious texts, and software manuals that are translated into several languages. Additionally, the World Wide Web represents a good and typical source for finding large-size and balanced parallel and monolingual text (Resnik and Smith, 2003).  However, many of the world languages do not have readily available digital corpora. Indigenous languages of the Americas do not have a web presence or text production comparable to richer resourced languages and it is difficult to find websites that offer their content the native languages. 

We explored the resources that are digitally available for some of the native languages spoken in the Americas. Regarding parallel corpora, the bible is a common source that contains translations to many of these languages, although it is not always straightforward to extract the content in a digital format. On the other hand, there are some projects that offer parallel content through a web search interface, e.g., Axolotl (Spanish-Nahuatl parallel corpus) that was mainly gathered from non-digital sources (books from several domains), the documents have dialectal, diachronic and orthographic variation~\cite{gutierrez2016axolotl}. Nahuatl is a Uto-Aztecan language spoken in Mexico (approx. 1.5M speakers) that lack of an orthographic normalization. In fact, this is the case for many languages spoken the Americas: large dialectal variation, and missing standardization. 


Also for the same language (Nahuatl), a comprehensive digital dictionary has been collected with information from five previous dictionaries~\cite{thouvenot2005gran}, these dictionaries date from 16th to 21st Century~\cite{olmos1547,walters2002diccionario}. The query interface of this resource is available online\footnote{\url{www.gdn.unam.mx}}.

Another language that belongs to the Uto-Aztecan family is Wixarika or Huichol (approx. 45K speakers). For this language, there is a parallel corpus (Wixarika-Spanish) that gathers translations of classic Hans Christian Andersen's literature~\cite{magerwixnlp}. In this case, the translations belong to one specific dialect (Nayarit). This resource can be fully downloaded from the web \footnote{\url{https://github.com/pywirrarika/wixarikacorpora}}.



The Shipibo-konibo language (approx. $26,000$ speakers) belongs to the Panoan language family, it is spoken in the Amazon region of Peru and Brazil. Several types of digital resources are available for this language. A parallel corpus between Spanish and this language was constructed using educational and religious documents \cite{galarreta:2017}. Moreover, Shipibo-konibo language has a POS tagged corpus, a set of words and its lemmas and an online-dictionary that has been recently released by~\newcite{pereira2017ship}.


Also spoken in South America, the Guarani language (approx. 8M speakers) belongs to the Tupi-Guarani family. \newcite{abdelali2006guarani} developed software for collecting Guarani resources from speakers, from this gathering they were able to construct a parallel corpus (Spanish-English-Guarani) and a monolingual corpus. There is also a digital Guarani dictionary \cite{ramirez1996} available online\footnote{\url{https://www.uni-mainz.de/cgi-bin/guarani2/dictionary.pl}}.

Quechua is one of the most spoken language families on the continent (approx. 9M speakers) but there is scarceness of corpora and language technologies. \newcite{espichan2017} released monolingual corpora in 16 Peruvian languages that belong to different linguistic families (including Quechua).

Regarding speech resources, Guarani has a spoken corpus comprised by $1,000$ phrases from $110$ different speakers, it was collected through a web interface~\cite{maldonado2016ene}, however, this dataset has not been publicly released. The Chatino language (approx. 45K speakers) is an Oto-Manguean language spoken in southern Mexico, recently a language documentation and revitalization project has been developed. They use automatic speech recognition and forced alignment tools to time align transcriptions. Parts of this resource are freely available \cite{CAVAR16.1006}.

There are some other types of datasets that are useful for developing language technologies, e.g., morphological annotated data. The CoNLL-SIGMORPHON 2017 Shared Task \cite{cotterell2017conll} released a large morphological database with inflection information of $52$ languages, including Haida ($7,040$ words), Navajo ($12,000$ words), and Quechua ($12,000$ words), all of these are indigenous languages spoken in the Americas. In the same way, there is a Oto-manguean inflectional class database which contains over $13,000$ verbal entries from twenty Oto-Manguean languages spoken in Mexico, along with information about each verb's inflectional class membership~\cite{palancaroto}. For morphological segmentation a data set of four Uto-Aztecan languages\footnote{Available at \url{http://turing.iimas.unam.mx/wix/mexseg}} were used and released by \newcite{kann2018} ($4,468$ segmented words from the Mexicanero, Nahutal, Wixarika and Yorem Nokki languages). The UQAILAUT Project contains roots, lexicalized words, infixes, noun and verb endings for the Inuktitut language\footnote{\url{http://www.inuktitutcomputing.ca/DataBase/info.php}} \cite{farley2012uqailaut}. Plain Cree language, spoken in North America (Algic language family) has also a lexicon databases ($16,452$ words) collected by \newcite{walters2002diccionario} and \newcite{wolfart1973computer}. 


To improve the data recollection, \newcite{dunham2014lingsync} developed tools for annotation of text and audio for  Blackfoot,  Gitksan,  Okanagan, Tlingit, Plains Cree,  Coeur d'Alene and Kwak'wala. With these tool they gather $19,187$ word forms, $324$ texts and $18.8$ GB of audio. 

A collection of datasets have been developed for the Mapudungun or Mapuche language spoken mainly in Chile (Araucanian language family) \cite{huenchullan2000data,monson2004data}. An audio dataset with $170$ hours of spoken Mapudungun, that covers three dialects ($120$ hours of Nguluche, $30$ hours of Lafkenche and $20$ hours of Pewenche) has been released. This resource contains a word list with the 70,000 most frequent full form words (stem plus inflections) to support a spelling checker for Mapudungun.  Also a bilingual Mapudungun-Spanish lexicon was included, containing sample entries $1,600$.

Annotated data corresponding to higher linguistic levels is harder to find. For instance, almost no treebanks have been developed for the indigenous languages of the Americas. To our knowledge, the only available dataset is a parallel aligned treebank between Quechua and Spanish~\cite{rios2008quechua} with $2,000$ annotated sentences.  

It is important to mention that many of the languages spoken in the Americas have a Wikipedia's set of articles available\footnote{The available languages in wikipedia can be consulted at: \url{https://es.wikipedia.org/wiki/Portal:Lenguas_indígenas_de_América}. Until the publication of this article there are only entries in: Nahuatl, Navajo, Guarani, Aymara, Klaalisut, Esquimal, Inukitut, Cherokee, and Cree.}. This is useful for building monolingual and comparable corpora. Furthermore, Wikipedia can be a helpful resource to predict Part-of-Speech (POS) tags for low resource languages and other tasks~\cite{hoenen2016wikipedia}. In any case, one common limitation of the digital resources for these languages is the lack of orthographic standardization and difficulties for processing certain types of characters. Table \ref{font-table} summarizes the above-mentioned resources.




\begin{table}[!t]
    \begin{center}
        \begin{tabular}{|p{3cm}|p{4cm}|p{3cm}|p{4.5cm}|}
            \hline \bf Type of resource & \bf Languages & \bf Size & \bf Reference\\ \hline
            Parallel& Nahuatl-Spanish & 18K sentences & \newcite{gutierrez2016axolotl}\\
            Parallel& Wikarika-Spanish & 8K sentences& \newcite{magerwixnlp}\\
            Parallel& Shipibo konibo - Spanish & 11.8K sentences & \newcite{galarreta:2017}\\
            Parallel& Spanish-English-Guarani & 250K sentences & \newcite{abdelali2006guarani}\\
            Parallel & 1259 languages &  & \newcite{mayer2014creating} \\
            POS Tagged & Shipibo konibo & 217 sentences& \newcite{pereira2017ship}\\
            Lemmatized words & Shipibo konibo & 3.5K words & \newcite{pereira2017ship}\\
            Dictionary & Shipibo konibo - Spanish & 3.5K words & \newcite{pereira2017ship}\\
            Dictionary & Nahuatl & & \newcite{palancaroto}\\
            Dictionary & Guarani & & \cite{ramirez1996}\\
            Speech & Guarani & 1K phrases & \newcite{maldonado2016ene}\\
            Speech & Chatino& 10 hours with Transcription & \newcite{CAVAR16.1006}\\
            Speech & Blackfoot,  Nata,  Gitksan,  Okanagan, Tlingit, Plains Cree, Ktunaxa, Coeur d’Alene, Kwak’wala& 19.8 GB & \newcite{dunham2014lingsync}\\
            Speech & Mapudungun  & 170 hours & \newcite{huenchullan2000data} and \newcite{monson2004data}\\
            Morphological Inflection& Quechua, Navajo, Haida & 31K words & \newcite{cotterell2017conll}\\
            Morphological Inflection & 20 Oto-Manguean languages & $13$K verbs & \newcite{palancaroto}\\
            Morphological Segmentation & Uto-Aztecan languages (Mexicanero, Nahutal, Wixarika, Yorem Nokki) &  4.4K words & \newcite{kann2018}\\
            Morphological segmentation & Inuktitut & 2K roots, 1.8K affixes & \newcite{farley2012uqailaut}\\
            Monolingual& 16 Peruvian languages & Unknown & \newcite{espichan2017} \\
            Monolingual& Plain Cree & $16$K words& \newcite{walters2002diccionario} and \cite{wolfart1973computer} \\
            Treebank&  Quechua & 2K sentences & \newcite{rios2008quechua}\\
            \hline
        \end{tabular}
    \end{center}
    \caption{\label{font-table} Digital available resources of American Indigenous Languages for NLP}
\end{table}


\section{Morphological segmentation and analyses}
\label{morphology:ref}

Morphology has been studied in NLP field focusing mainly on the following tasks: lemmatization, stemming, segmentation, analysis and inflection/reinfection. These tasks serve to other higher level tasks such as machine translation. On this regard, there have been several studies which have been applied to the Americas languages.

In NLP, lemmatization and stemming methods are used to reduce the morphological variation by converting words forms to a standard form, i.e., a lemma or a stem. However, most of these technologies are focused in a reduced set of languages. For languages like English there are plenty of resources, however, this is not the case for all the languages. Specially for languages in the Americas with rich morphological phenomena, and not always suffixal, where it is not enough to remove inflectional endings in order to obtain a stem.

Morphological segmentation is the task of splitting a word into the surface forms of its smallest meaning-bearing units, its \emph{morphemes}. On the other hand, Morphological analysis not only focuses in the segmentation of words, but also in assigning tags to each part of the word. There are several approaches to do these tasks, i.e., rule-based, semi-supervised  and unsupervised \cite{Goldsmith:2001:ULM:972667.972668,creutz-lagus:2002:ACL02-MPL,Kohonen:2010:SLC:1870478.1870488}. Some examples of rule-based methods applied to the Americas languages are the Finite State approaches to model the morphology of a language:  plains Cree \cite{arppe2017converting,harrigan2017learning,wolfart1973computer,snoek2014modeling}, East Cree~\cite{arppe2017converting} , for the East Odawa dialect of Ojibwe~\cite{bowers2017morphological}, for Mohawk (Iroquoian language family)~\cite{assini2013natural}, for the Bribri (Chibchense language family)~\cite{solorzano2017} using the FOMA tool~\cite{hulden2009foma}, Quechua~\cite{vilca2012analizador,monson2006building}, Mapudungun~\cite{monson2006building}, and the Argentinian branch of Quechua and Toba~\cite{porta2010use}. More recently a new hybrid approach of finite-state transducer (FST) with statistical inference is part of the \textit{Basic Language Technology Toolkit for Quechua}~\cite{rios2016basic}. 

For Uto-Aztecan languages, there exists a computational tool called ``\emph{chachalaca}'' that performs morphological analysis~\cite{Marc2} of Nahuatl. This is a rule-based software focused on Classical Nahuatl, it is able to generate more than one morphological analysis candidate per word. It is based on grammars that describe most of the 16th-century-word constructions. Additionally, \newcite{magerwixnlp} propose a morphological segmentation tool for the Wixarika language, with a supervised approach, using previous given morphological tables and a probabilistic model to infer the inherent morphological rules. 

Regarding to unsupervised methods, neural methods have been used to tackle the rich morphology of the languages of the continent. \newcite{micher2017improving} propose a Segmental Recurrent Neural Network (RNN) for segmenting and tagging Inuktitut. \newcite{kann2018} used a set of extensions to the Encoder-Decoder RNN architecture with Gated Recurrent Units (GRU) for automatically segmenting four Uto-Aztecan languages (Mexicanero, Nahuatl, Wixarika and Yorem Nokki). Semisupervised segmentation approaches like Morfessor have also been successfully applied to Nahuatl~\cite{ximena2017bilingual}. For the Uto-Aztecan language Tarahumara and the Mayan language Chuj, there are works that try to automatically discover affixes through unsupervised approaches \cite{medina2007affix,medina2008affix,medina2006experimento}.

Lately, there has been interest in the reinflection task, i.e., generating an inflected form for a given target tag and lemma. The CoNLL-SIGMORPHON Shared Task \cite{cotterell2016sigmorphon,cotterell2017conll} released a dataset for reinflection of 52 languages, including 3 Native American languages. The systems that got the best performance \cite{kann2016med,kann2017lmu,makarov2017align}.

In some cases it is difficult to disambiguate between homonym morphs. To deal with this problem, \newcite{rios2008quechua} used Conditional Random Fields (CRFs)~\cite{lafferty2001conditional}.

Most of the methods that we found that deal with morphology are based on FST. However, the indigenous languages of the continent are far too diverse, it would be expensive to build such analyzers with expert knowledge for each language, besides the fact that the analyzers need to be constantly updated to cope with language change~\cite{creutz2005unsupervised}. 

\section{Machine Translation}
\label{mt:ref}

Machine Translation is a natural task for indigenous languages, since it might provide a communication window with more popular languages. The development of MT systems for indigenous languages have follow the trends in the field, from rule-based, to statistical and neural based approaches. 

 Rule-Based Machine Translation (RBMT) approaches are sometimes suitable for low resource languages since they do not require aligned parallel corpora. However. In recent years, research on data-driven approaches has increased, with the aim to overcome the scarcity of data using different methods. In any case, translation of low-resource languages represents an interesting and active research problem in the NLP field. 

In the case of native American languages, there have been some efforts in building rule-based systems. The Apertium system~\cite{tyers2009apertium,forcada2011apertium} is a big help for this approach, and at least two languages has translation teams working with it. This is the case of Quechua (Eastern Apurimac Quechua and Cusco Quechua)-Spanish~\cite{caverotraductor} and Spanish-Wayuunaiki  (Spoken in Venezuela and Colombia)~\cite{fernandez2013design}. Other RBMT systems were created for Aymara-Spanish (spoken in Peru)~\cite{coler20144} Wayuunaki-Spanish, Quechua-Spanish and Mapuche-Spanish~\cite{monson2006building}. For the latter a web available translator\footnote{\url{http://142.4.219.173/wt/}}\cite{gonzalez:2016} is available. We found that there are mobile apps for translating indigenous languages, this is the case of Zapotec-Spanish language pair (spoken in Mexico from the Oto-Manguean family)\footnote{\url{https://play.google.com/store/apps/details?id=com.SimplesoftMx.Didxazapp&hl=es}}.

All of this RBMT systems have a set of shortcomings. The majority is not able to translate complex constructions, specially when the languages are distant from each other, which increases the complexity of the machine translation rules. One way to overcome this is to include linguistic information, e.g., morphology, syntax. However, this kind of knowledge or linguistic tools is not always available, especially for low-resource languages.  Experiments using the Example Based Machine Translation (EBMT) methodology are not common,  we only found the work of~\newcite{llitjos2005building} and \newcite{monson2006building} for the Mapuche-Spanish pair.

Statistical Machine Translation (SMT) Systems are data-driven since they use vast amounts of parallel corpora to model the translations between sentences or subunits. Their performance is highly dependent on the number of training data; they represent a challenge when low resource conditions are faced. In the case of the native languages of the Americas, they tend to be morphologically rich and this must be taken into account to improve the translation and reduce the data sparseness. An example of this is the Wixarika-Spanish SMT system that aligns Wixarika morphemes with Spanish words or tokens \cite{mager:2016,mager:2017}\footnote{Available at \url{http://turing.iimas.unam.mx/wix}}. A similar case can be found for the Nahuatl-Spanish pair. Uto-Aztecan languages can be highly agglutinative with the polysynthetic tendency, \newcite{gutierrez2015bilingual} extracts bilingual correspondences from a parallel corpus, by aligning the Nahuatl non-grammatical morphs to Spanish words. Another example was collected for the pair Mixteco-Spanish~\cite{santiago2017}. The same trend can be observed in SMT for Shipibo-konibo \cite{galarreta:2017}.

Regarding commercial systems, Microsoft has targeted some languages spoken in Mexico, Mayan and Otomi (Queretaro variant)\footnote{\url{https://www.microsoft.com/en-us/translator/languages.aspx}}. SMT was also applied for the Guarani, it translates to Spanish, but also to English, French, Italian, German and Portuguese\footnote{\url{http://www.iguarani.com/?p=traductor}}.

Recently, there has been an increasing interest in Neural Machine Translation (NMT) models, which are also statistical based, but they use neural networks architectures that are feed with very big amounts of parallel corpora. \newcite{mager2018dh} showed that in such low-resource scenarios, translating from Mexicanero, Nahuatl, Purepecha, Wixarika and Yorem-Nokki to Spanish, SMT systems achieve better performance than NMT.  Even though these architectures are not suitable for low-resource settings, there have been some recent efforts to adapt them. \newcite{Soriano2018} experimented with the Mexican Purepecha (an isolated language with about 140 thousand speakers) using the OpenNMT toolkit \cite{2017opennmt}. \newcite{tiedemann2018emerging} took the massive bible corpus~\cite{mayer2014creating} and trained a multilingual NMT model to improve overall translation performance. Experiments included Oto-manguean, Quechua and Mayan families.  
Moreover, empirical results~\cite{vania-lopez:2017:Long} show that problem of data sparsity of rich morphological languages can be handled with subword models: the usage of character level NMT improve performance over token level translation and unsupervised morphological segmentation~\cite{creutz-lagus:2002:ACL02-MPL}. But their experiment also conclude that a canonical segmentation enhances character level translation. 

In order to alleviate the lack of resources automatic data recollection has been proposed, this has been tried for Guarani language \cite{abdelali2006guarani}. Moreover, it would be very useful to have big repositories of translated texts. One alternative is to create parallel corpora between many languages using manual translations (controlled elicitation) as described for the Mapudungun (Mapuche) language~\cite{probst2001design}.

In any case, SMT and NMT systems should be adapted to deal with the scarcity, of the sparseness of word forms and the rich morphology of languages. Although there are works that try to deal with morphology \cite{virpioja2007morphology,popovic2004towards,costa2016character,sennrich2016neural,dalvi2017understanding}, they are rarely applied to Native American languages.
\section{Other studies and tools}  
\label{other:ref}
\subsection{Multilinguality and Code-Switching}  
\label{code-swiching:ref}

Most native speakers of indigenous languages are at least bilingual, they have to communicate using the primary or official language of their own country, i.e., Spanish, Portuguese, French or English. Only few communities remain completely monolingual in their native language, moreover, modern migrations and the use of social networks contribute to bilingualism and code-switching.

Code-switching occurs when a speaker alternates between two or more languages in a conversation. This adds a challenge when doing NLP for this kind of data. Code-switching is not a new phenomenon, it can be found in historical documents, \newcite{garrette2016unsupervised} proposes an unsupervised approach of paired encoding (words and characters) to improve language modeling (Latin, Spanish and Nahuatl) in an Optical character recognition (OCR) task. \newcite{king2013labeling} applies weakly supervised methods for labeling the language of each word in documents that can have many mixed languages. The targeted languages are 30, including Chippewa, Nahuatl, and Ojibwa.

Being able to automatically detect Code-switching could be useful when doing NLP for minority languages, for instance to use the web as a source for a corpus.


From the quantitative linguistics perspective, parallel corpora of an outstanding number of languages have been extracted from the Bible and used to perform typological studies in many languages, included native languages of the Americas. For instance, exploring the tense behavior \cite{asgari2017past}, contrasting the morphological complexity in many languages \cite{bentz2016comparison,kettunen2014can} just to mention some. 


\subsection{Language Tools}
\label{other-tasks:ref}
For some rich resource languages, there are already available NLP tools that deal with several phenomena and linguistics levels of processing. However, for low resource languages, there is still much work to do. In this section we summarize some of the works related with POS Tagging, OCR, Parsing, Spell Checking, Language Identification and other tasks, that we have found for the languages of the Americas.

\textbf{Speech synthesis and recognition} has made some progress for the Raramuri language~\cite{urreatowards}, using a unit selection approach based on function words, suffix sequences and diphones of the language. For speech recognition, \newcite{maldonado2016ene} applied on Guarani a Hidden Markov Model (HMM) with the CMU Sphinx toolkit~\cite{lamere2003cmu}. \newcite{coto2016alineacion} proposes an automatic aligner of text and voice for the indigenous language Bribri of Costa Rica. 

\textbf{Part-of-Speech (POS) tagging} assigns a category from a given symbol set to each token in an input string. It is used as a prepossessing step that serves as input for other tasks or for higher level NLP task. POS Tagging was also incorporated into the Peruvian Shipibo-konibo NLP toolkit~\cite{pereira2017ship}.





\textbf{Spell checking} is not a trivial task for highly agglutinative and polysynthetic languages, that can't rely on a token based evaluation, and need sub-word level models. We found only two tools that handle this issue: \newcite{monson2006building} build a dictionary based tool for Quechua and \newcite{alva2017spell} for Shipibo-Konibo used rule-based analysis and dictionaries. 

Another field that is crucial to increase the amount of digitized data for other tasks is \textbf{Optic Character Recognizing} (OCR). \newcite{maxwell2017endangered} studied the challenges regarding the digitalization of Tzeltal-Spanish, Muinane-Spanish,  Cubeo-Spanish dictionaries. \newcite{garrette2016unsupervised} developed an unsupervised transcription model for dealing with orthographic variation in digitized historical documents, some of them were written in Nahuatl.

In the context of indigenous \textbf{Language Identification} (LID), \newcite{espichan2017} studied LID on $17$ languages from the Arawak, Aru, J\'{i}baro, Pano and Quechua linguistic families. The proposed models were flexible enough to handle the lack of orthographic standardization of the language.
%

Another important NLP task is \textbf{parsing}.   The research in this area is also weak, however, the Quechua-Spanish Treebank helped to perform some experiments in this topic~\cite{bresnan2015lexical,rios2016basic}. For other languages parsing experiments were performed on Ayamara \cite{homola2011parsing} using Lexical-Functional Grammar (LFG). 

\section{Discussion}
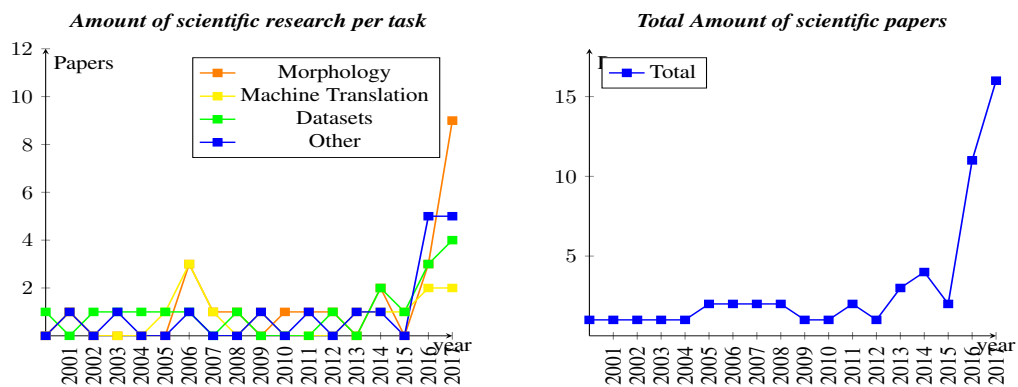
\begin{figure*}[h!]
\begin{minipage}{0.49\textwidth}
\centering
\begin{adjustbox}{width=.80\textwidth, height=.65\textwidth,}
\begin{tikzpicture}
\begin{axis}[legend pos=north east,
axis lines=middle,
ymin=0,
x label style={at={(current axis.right of origin)},anchor=north, below=0mm},
    title={\textit{\textbf{Amount of scientific research per task}}},
    xlabel=year,
    ylabel=Papers,
  xticklabel style = {rotate=90,anchor=east},
 enlargelimits = false,
  xticklabels from table={workpyear.dat}{X},xtick=data,
  ymin=0,ymax=12]
\addplot[orange,thick,mark=square*] table [y=Morphology,x=X]{workpyear.dat};
\addlegendentry{Morphology}
\addplot[yellow,thick,mark=square*] table [y=MT,x=X]{workpyear.dat};
\addlegendentry{Machine Translation}
\addplot[green,thick,mark=square*] table [y=DS,x=X]{workpyear.dat};
\addlegendentry{Datasets}
  \addplot[blue,thick,mark=square*] table [y=Other,x=X]{workpyear.dat};
  \addlegendentry{Other}]
\end{axis}
\end{tikzpicture}
\end{adjustbox}
\end{minipage}
\begin{minipage}{0.49\textwidth}
\begin{adjustbox}{width=.80\textwidth, height=.65\textwidth,}
\begin{tikzpicture}
\begin{axis}[legend pos=north west,
axis lines=middle,
ymin=0,
x label style={at={(current axis.right of origin)},anchor=north, below=0mm},
    title={\textit{\textbf{Total Amount of scientific papers}}},
    xlabel=year,
    ylabel=Papers,
  xticklabel style = {rotate=90,anchor=east},
 enlargelimits = false,
  xticklabels from table={totalworkpyear.dat}{X},xtick=data,
  ymin=0,ymax=18]
  \addplot[blue,thick,mark=square*] table [y=all,x=X]{totalworkpyear.dat};
  \addlegendentry{Total}]
\end{axis}
\end{tikzpicture}
\end{adjustbox}
\end{minipage}
    \caption{NLP papers and digital resources that contain any indigenous language of the Americas (between 2000 and 2017)}
\label{fig:research}
\end{figure*}

\noindent
Study of the languages of the Americas has increased in the recent years, in both the linguistic and the language technology fields. Many factors have contributed to this, such as speakers self-awareness about the importance of their languages and digital inclusion. Figure~\ref{fig:research} shows that many of the papers that we reviewed, were published from year 2000 to present day, with a notable increase in the activity in the last five years. The most studied NLP tasks are Machine Translation and Morphology, however, from 2013 upon now, other tasks, e.g., POS-tagging, parsing, speech, spell correction, also received attention.

Despite the fact that we found NLP contributions for around $35$ languages, \manuel{Falta actualizar el dato} this is still a small number if we take into account the big diversity and number of languages that exist in the continent. Table~\ref{families} showed that some linguistic families have concentrated the attention, but even for these languages the developed technology is not enough. We noticed that North American languages are the most studied, despite some of them don't have a big number of speakers compared to other indigenous languages, e.g., Navajo, Haida, Cree, Chippewa, Ojibwa, Blackfoot, Nata, Gitksan, Okanagan, Tlingit, Plains Cree, Ktunaxa, Coeur d’Alene, Kwak’wala, and Inuktitut. Uto-Aztecan language family that includes languages like  Raramuri, Nahuatl, Wixarika, Yorem Nokki, Mexicanero (spoken mainly in Mexico) have also received attention from the NLP community. Regarding to South America, the most spoken native languages, Quechua, Mapuche, Guarani and Ayamara, have several resources available. Surprisingly, languages with less speakers such as Shipibo-konibo,  Arawak, Aru, J\'{i}baro, Pano and Wayuunaki have been also studied.

The diversity in the linguistic phenomena of these languages makes developing language technologies a challenging task. In recent years, NLP and Machine Learning fields have paid attention to low resource settings, organizing workshops and special tracks to tackle this issue. Indigenous languages could be greatly benefit from this kind of research in the future. In particular, American languages with rich morphology, e.g., agglutinative and polysyntehthic, seems to benefit from approaches that take into account the morphology and sub-word modeling.

We also noticed that some NLP tasks that are considered almost solved for languages like English, need to be adapted or started from scratch when applied to the languages of the American continent. Moreover, fields like machine translation could enable in the future the creation of multilingual technologies for all of the languages in the world that face a similar situation, this could have a great impact in these communities.

\section{Conclusions}
\label{conclusions:ref}
In this work, we presented a review of NLP research focused on Indigenous Languages of the Americas. We showed which languages have available digital resources and their related tools. Research has focused in tasks like morphology and machine translation, however, there is still a lot of work to be done since these languages exhibit a wide range of linguistic phenomena while resources are scarce. 

Through this work, we discussed some of the challenges that must be taken into account, e.g., small datasets, high dialectal variation, rich morphology, lack of orthographic normalization, scarcity of linguistic preprocessing tools. 

NLP research for these languages can broad the understanding of human language structures and help to build more general computational models. Moreover, the development of language technologies can have a positive social impact for the speakers of the indigenous languages, helping to maintain the living cultural heritage that each language represents.

\section*{Acknowledgements}
This work was supported by the Mexican Council of Science and Technology (CONACYT), fund 2016-01-2225. We will also thank the reviewers for their valuable comments and to Katharina Kann for her comments and support. 

\bibliographystyle{acl}
\bibliography{acl2016}
\newpage
\section*{Appendix A. Language Families}
\label{appendix1}

Table \ref{families} summarizes the language families for which we were able to identify some digital and technological resources during this research. We distinguish among only two geographical macroareas: North America (it includes Central America) and South America \cite{wals}. 

\begin{table}[h]
    \begin{center}
        \begin{tabular}{|c|c|r||c|c|r|}
            \hline \bf L. Family & \bf Macroarea & \bf Papers & \bf L. Family & \bf Macroarea & \bf Papers\\\hline
Uto-Aztecan & North A.& 16&Mayan&	North A. & 4\\
Oto-Manguean&	North A.&3&Arawakan&	South A.&3\\
Haida&	North A.&4&Ayamaran&	South A.&2\\
Na-Dene&	North A.&5&Aru&	South A.&1\\
Eskimo-Aleut&	North A.&2&Jibaro&	South A.&1\\
Algic&	North A.&8&Bora–Witóto &	South A.&1\\
Tsimshianic&	North A.&1&Tucanoan &	South A.&1\\
Penutian&	North A.&1&Araucanian&	South A.&7\\
Salishan&	North A.&2&Panoan &  South A.& 4\\
Wakashan&	North A.&1&Tupian & South A.&5\\
Iroquoian&	North A.&1&Quechuan &South A.&15\\
Chibchan&	North A.&2&Guaicuruan &South A.&1\\
\hline
        \end{tabular}
    \end{center}
    \caption{\label{families}Language families (L. Family) for which some technology was found, and the number of NLP/Computer Linguistic papers referring to each (one paper can reference more than one languages). North A. stands for North America and South A. for South America.}
\end{table}

\end{document}